\documentclass[10pt,twocolumn,letterpaper]{article}

\usepackage{iccv}
\usepackage{times}
\usepackage{epsfig}
\usepackage{subcaption}
\usepackage{graphicx}
\usepackage{amsmath}
\usepackage{amssymb}
\usepackage{algorithm}
\usepackage[noend]{algpseudocode}
\usepackage{booktabs, makecell}
\usepackage{multirow}
\usepackage{lipsum}

\usepackage{array} 
\newcolumntype{C}{>{\centering\arraybackslash}p{\dimexpr.5\wexp-\tabcolsep}}

\usepackage[numbers]{natbib}

\usepackage[pagebackref=true,breaklinks=true,letterpaper=true,colorlinks,bookmarks=false]{hyperref}

\iccvfinalcopy 

\def\iccvPaperID{38} 

\ificcvfinal\fi

\begin{document}

\title{Improving Replay Sample Selection and Storage for Less Forgetting in Continual Learning}

\author{Daniel Brignac\\
University of Arizona\\
Tucson, Arizona\\
{\tt\small dbrignac@arizona.edu}
\and
Niels Lobo\\
University of Central Florida\\
Orlando, Florida\\
{\tt\small niels@cs.ucf.edu}
\and
Abhijit Mahalanobis\\
University of Arizona\\
Tucson, Arizona\\
{\tt\small amahalan@arizona.edu}
}

\maketitle
\ificcvfinal\thispagestyle{empty}\fi

\begin{abstract}
Continual learning seeks to enable deep learners to train on a series of tasks of unknown length without suffering from the catastrophic forgetting of previous tasks. One effective solution is replay, which involves storing few previous experiences in memory and replaying them when learning the current task. However, there is still room for improvement when it comes to selecting the most informative samples for storage and determining the optimal number of samples to be stored. This study aims to address these issues with a novel comparison of the commonly used reservoir sampling to various alternative population strategies and providing a novel detailed analysis of how to find the optimal number of stored samples.


\end{abstract}

\section{Introduction}

Deep learning has revolutionized the field of computer vision, achieving human-like capabilities of image understanding and perception. Unlike humans, however, deep learners consistently struggle to adopt new knowledge while maintaining performance on previously learned tasks. This is the problem known as catastrophic forgetting~\cite{cat_forget} in which a learner’s performance significantly diminishes as it acquires knowledge for new tasks. This motivates the study of continual learning~\cite{cl_RING, cl_Thrun} to address the problem of catastrophic forgetting.

In continual learning, a learner is presented with a sequence of tasks of unknown length where the only data available is that of the current task at hand. As new tasks arrive, we wish to learn each new task while preserving the knowledge learned from previous tasks. During inference, we may be presented with data from any of the previously learned tasks, thus the retention of previous knowledge is imperative when adapting to new tasks. 

The inference stage primarily comes in three flavors: task-incremental learning (task-IL), domain-incremental learning (domain-IL) and class-incremental learning (class-IL) \cite{CL_scenarios}. Task-IL and domain-IL are generally considered the easier scenarios as we are either given the task-ID at test time in task-IL, or we must only solve for the current task at hand in domain-IL. Class-IL is significantly more challenging as we must infer for all tasks seen so far without the revealing of a task-ID. Because of this, class-IL has become a primary focus of recent continual learning works \cite{Lipschtiz, dualnet, ER-ACE, example_influence}.

Replay~\cite{iCaRL, ER, DER} is a commonly used approach to remedy the problem of catastrophic forgetting in both class-IL and task-IL. The concept of replay draws inspiration from Complementary Learning Systems theory of humans which posits that recent experiences stored in the hippocampus develop connections to the neocortex that become ingrained over time to eventually be encoded in long-term memory \cite{CLS, CLS_Replay}. As such, when replay is employed in artificial learners, a small amount of previously learned data is stored in memory to then be “replayed” during the training of the current task to emphasize the previously encoded connections of the learner and thus avoid forgetting.

When using replay methods, it is important to have a small canonical set of exemplars stored in memory that effectively capture the underlying class distributions of the dataset. Thus, the selection of \emph{which samples to store} is a non-trivial task. Previous replay methods rely on reservoir sampling~\cite{ER, DER, MIR, ER-ACE, GSS} to populate memory with past data. As reservoir sampling is a random sampling method, this could lead to the storage of redundant and potentially insignificantly informative data points in memory causing replay methods to not perform to their maximal capability as recently demonstrated in ~\cite{OnlineMemSelection, OnlineCoresetSelection}.

There is also no consensus regarding \emph{how many samples should be stored} to be used for replay. Naturally, as we store more samples in memory, we expect performance to increase, however, as this number grows, we start to deviate from the constraints of continual learning where we seek to store minimal information. There must exist some small optimal number of samples to be stored to make maximum use of replay methods.

In this work we study the two questions \emph{which samples to store}, and \emph{how many samples should be stored} in memory by comparing the commonly used approach of reservoir sampling to three other memory population strategies. We show through extensive empirical evaluation that reservoir sampling leads to greater forgetting when compared with more strategic population approaches. We additionally detail two methods to address the question of how many samples should be stored based on an analysis of significant eigenvectors and eigenvalues. We proceed to show that memory populated according to these two criteria leads to overall more competitiveness and better performance of all population strategies.

\section{Related Works}

Continual learning methods can generally be grouped into three categories: (i) regularization methods, (ii) architectural methods and (iii) replay methods. Regularization methods introduce a new loss term meant to penalize significant drift from previously learned parameters \cite{EWC, LwF, LFL, HyperNets}. Architectural methods seek to adapt existing network architectures such that various portions of the model contain global and/or shared knowledge while others contain task-specific knowledge \cite{IBPWF, DynExp, CalCNNs, ProgNets, dualnet}. Historically, regularization methods and architectural methods underperform when compared directly to replay methods further motivating the study of and improvements to replay.

In replay methods, we are allowed to store some small subset of data in a memory buffer and selectively “replay” samples from this buffer when training on the current task to maintain previous task performance. Earlier works include \cite{GEM} and \cite{AGEM}, both of which use the memory buffer as a constraint on gradient updates to ensure that loss remains low on the buffer samples. These gradient-based methods for replay in general show poor performance when compared to experience replay methods \cite{iCaRL, ER, DER, MIR, HAL, co2l, ER-ACE}. In experience replay, it is common practice to store any of the raw data sample, the label, the logits, or a combination of all three. All the the aforementioned experience replay methods select the samples to be stored by some random sampling method, such as reservoir sampling or uniform sampling, leading to the potential storage of insignificant data. 

A number of previous works suggest to either populate the memory buffer with some fixed arbitrary number of samples as new classes are encountered or to empirically find optimal buffer size for a desired task performance \cite{MIR, 1995_rehersal_methods}. These methods for expanding the buffer are rooted in heuristics and thus not necessarily optimal further motivating the study for optimal growth of the buffer for learning new tasks.

\section{Methodology}

\setlength{\textfloatsep}{10pt}
\renewcommand{\algorithmicrequire}{\textbf{Input:}}
\begin{algorithm}
    \caption{Memory Buffer Population} \label{alg: buffer}
    \begin{algorithmic}
    \Require $\mathcal{T}$, $\mathcal{D}_t$, $\mathcal{M}$, buffer size (if not dynamic), populate $\in$ \{\emph{reservoir, herding, GSS, IPM}\}, training-strategy\\
    
    \State $\mathcal{M} \gets \emptyset$
    \For{$t$ in $\mathcal{T}$}
        \For{\emph{minibatch} in $\mathcal{D}_t$}
            \If{$\mathcal{M} = \emptyset$}
                \State \emph{train}($\mathcal{D}_t$) according to training-strategy
                \If{populate $\in$ \{\emph{resrvoir, GSS}\},}
                    \State $\mathcal{M} \gets \text{populate}(minibatch, \text{ buffer size})$
                \EndIf
            \Else
                \State \emph{train}($\mathcal{D}_t \cup \mathcal{M}$) according to training-strategy
                \If{populate $\in$ \{\emph{resrvoir, GSS}\},}
                    \State $\mathcal{M} \gets \text{populate}(minibatch, \text{ buffer size})$
                \EndIf
            \EndIf
        \EndFor
    \If{populate $\in$ \{\emph{herding, IPM}\},}
        \State $\mathcal{M} \gets \text{populate}(\mathcal{D}_t, \text{ buffer size}$)
    \EndIf
    \EndFor
    \end{algorithmic}
\end{algorithm}

Consider a sequence of $\mathcal{T}$ tasks where each task has an associated dataset $\mathcal{D}_t=\{(x_i^t, y_i^t)\}$ for each $t \in \{1,2,...,\mathcal{T}\}$ and $i=1,..., N_t$ where $x_i^t$ denotes the  $i^{th}$ sample of the $t^{th}$ task, $y_i^t$ its associated ground truth label, and $N_t$ the number of samples in task $t$. We assume each task $t$ contains a unique, non-overlapping set of classes drawn from an i.i.d distribution. In our continual learning formulation, we seek to sequentially learn each $\mathcal{D}_t$ in an offline setting while maintaining performance on every $\mathcal{D}_k$ for $k<t$ by employing a memory buffer $\mathcal{M}$ which is populated according to a specific population strategy. For $t=1$, we train only on $\mathcal{D}_t$ and for $t>1$, we train on $\mathcal{D}_t \cup \mathcal{M}$.

Since we are primarily concerned with strategic buffer population, the training method in which we perform continual learning is agnostic of the memory buffer population. Thus, we can use any training strategy along with any buffer population strategy studied herein. We give an overview of each population strategy studied and then present a novel scheme to identify how many samples per class should be stored in $\mathcal{M}$.

We make a note of the distinction between fixed memory buffers and dynamic memory buffers. A fixed memory buffer refers to a buffer $\mathcal{M}$ that is fixed in the amount of data it can hold (\eg, a fixed buffer size of 200 can contain a maximum of 200 data samples) while a dynamic buffer may grow in size as new classes are encountered. We denote the size of the fixed buffer as $|\mathcal{M}|$. Traditionally, fixed memory buffers are used whenever replay methods are employed. To the best of 
our knowledge, we are the first to consider dynamic memory buffers as discussed in Section \ref{sec:dynamic_buffers}. When the buffer is not described to be either fixed or dynamic, the method is agnostic of buffer type. An overview of our described approach is given in Algorithm \ref{alg: buffer}.

\subsection{Reservoir Sampling}
In reservoir sampling \cite{reservoir}, we are presented with a stream of data in which we randomly sample from the stream and store each sample in $\mathcal{M}$. If a sample is selected to be stored when $\mathcal{M}$ is saturated, we then randomly replace a sample in $\mathcal{M}$ with the current selected sample to be stored. 

In practice, reservoir sampling tends to favor the storage of samples from earlier encountered tasks. This causes the minimal storage of samples from downstream tasks and an overall unbalanced fixed buffer leading to greater forgetting, particularly when the fixed buffer size is small.

In addition to favoring the storage of earlier encountered task data, reservoir sampling has no mechanism to differentiate between whether a selected sample to be stored is informative or redundant. This leads to the potential storage of insignificant data which in turn can diminish the network's previously learned decision boundaries \cite{Lipschtiz}. This motivates the study of memory buffer population strategies that can always store the next most informative sample and mitigate forgetting.

\subsection{The Herding Algorithm}
A natural first choice of substitute to reservoir sampling is the herding algorithm as proposed in iCaRL \cite{iCaRL} which is an extension of Welling's herding in \cite{herding}. Herding seeks to store samples that best represent the sample's class mean in feature space. In this sense, the herding algorithm can be thought of as a greedy mean preserving scheme as each selected sample is the closest to its learned class mean.

Formally, for a learned class mean $\mu_c$ in $D$-dimensional feature space and a mapping from image space to feature space given by $\phi:x_i^{t,c} \rightarrow \mathbb{R}^D$ where $c$ denotes the class label of image $x_i^t$, herding seeks to find the sample $x_i^{t,c}$ to add to the class exemplar set $P_c$ in memory buffer $\mathcal{M}$ that minimizes the distance to $\mu_c$ as
\begin{equation}
    \label{eq:herding}
    \underset{x_i^{t,c}}{\operatorname{argmin}} \left| \left|\mu_c - \frac{1}{K} \left( \phi(x_i^{t,c}) + \sum_{j=1}^{K-1} \phi(p_j) \right)\right|\right |_2
\end{equation}
where $K$ denotes the number of samples to be stored for class $c$ and $p_j$ denotes a sample $j$ belonging to $P_c$.

Because herding relies on the well learned features for each class to store samples, herding should be performed at the conclusion of each task. This makes herding most suitable for the offline continual learning scenario in which we are allowed multiple iterations through $\mathcal{D}_t$ before moving to the next subsequent task.

\subsection{Gradient Sample Selection}
\label{sec: GSS_sec}

Gradient sample selection (GSS) is taken from \cite{GSS} where the objective is to optimize the loss on current task data while maintaining minimal loss on previous task data. To ensure that loss remains low on previous tasks, GSS seeks to maximize the diversity of the buffer $\mathcal{M}$ by selecting samples whose gradient angle is maximal compared to all other samples currently stored (\ie, maximizing gradient direction variance). This formulation involves solving a quadratic integer programming of polynomial complexity w.r.t. the buffer size thus, a greedy formulation of GSS is used instead.

In the greedy method, GSS maintains a score $r_i$ for each sample $i$ in the buffer based on the maximal cosine similarity of the current sample with the other samples in the buffer given by
\begin{equation}
    \label{eq:GSS}
    r_i = \underset{i}{\operatorname{max}} \frac{\langle g_i, G \rangle}{||g_i||_2 ||G||_2}
\end{equation}
where $g_i$ and $G$ denote the gradients of the current sample $i$ and the set of samples stored in $\mathcal{M}$ respectively. 

While GSS aims to select and store samples based on maintaining maximum variance of gradient direction, there is still a randomness component in determining samples to potentially be replaced. When a sample is selected and deemed appropriate for replacement, there is no guarantee that this is a least informative sample, as determined by GSS, and could thus potentially lead to greater forgetting. In addition, GSS has no mechanism for class balancing as samples are randomly added and replaced which can further hinder performance.

\subsection{Iterative Projection and Matching}
Iterative projection and matching (IPM) \cite{IPM} is drawn from active learning in which we seek to find the most informative data points for training. Here, we adopt IPM to select the most informative data points for storage in $\mathcal{M}$ to address the shortcomings with reservoir sampling while also maintaining a balanced buffer similar to the strategy used in herding.

Let $\boldsymbol{A}_c \in \mathbb{R}^{N_c \times D}$ denote the matrix of features $\phi(x_1^t),...,\phi(x_{N_c}^t) \in \mathbb{R}^D$ where $N_c$ denotes the number of samples in class $c$, $D$ the dimension of features, and $\phi: \mathbb{R}^{H \times W \times C} \rightarrow \mathbb{R}^D$ a mapping from image space to feature space (\ie, the learned network). The $n^{th}$ row in $\boldsymbol{A}_c$ is formed by $\phi(x_{n_c}^t)^T$.  We seek to reduce $\boldsymbol{A}_c$ to some $\boldsymbol{A}_R \in \mathbb{R}^{K \times D}$ where $K$ is the number of samples to be stored from class $c$.

Let $T \subset \{1,...,N_c\}$ with $|T|=K$ be the set of selected samples. Then, we can project the rows of $\boldsymbol{A}_c$ onto the span of the selected rows indexed by $T$. We denote this operation as $\boldsymbol{\omega}_T(\boldsymbol{A}_c)$. As done in \cite{SMRS}, we can now cast the replay sample selection problem as the optimization problem 
\begin{equation}
    \label{eq:optim}
    \underset{|T|=K}{\operatorname{argmin}} ||\boldsymbol{A}_c - \boldsymbol{\omega}_T(\boldsymbol{A}_c)||_F^2
\end{equation}
where $||\cdot||_F$ is the Frobenius norm. This problem is NP-Hard however as we must search all subsets $T$ over $\boldsymbol{A}_c$~\cite{ColSubSelect}. Thus, we use the IPM \cite{IPM} algorithm to approximate \ref{eq:optim}.

We can express $\boldsymbol{\omega}_T(\boldsymbol{A}_c)$ as a rank-$K$ factorization $\boldsymbol{U}\boldsymbol{V}^T$ where $\boldsymbol{U} \in \mathbb{R}^{N_c \times K}$ and $\boldsymbol{V}^T \in \mathbb{R}^{K \times D}$ and modify \ref{eq:optim} by recasting it as two sub-problems \cite{IPM, SP_KSP}
\begin{equation}
    (u,v) = \underset{u,v}{\operatorname{argmin}} ||\boldsymbol{A}_c-\boldsymbol{u}\boldsymbol{v}^T||_F^2 \text{ s.t. } ||\boldsymbol{v}|| = 1
\end{equation}
\begin{equation}
    m^{(1)} = \underset{m}{\operatorname{argmax}} |\boldsymbol{v}^T\boldsymbol{\rho}|
\end{equation}
where $\boldsymbol{\rho} = \phi(x_{n_c}^t) / ||\phi(x_{n_c}^t)||_2$, $m^{(1)}$ is the index of the first selected data point, and $(x_i^t,y_i^t)_{m^{(1)}}$ is the selected point to be stored in memory.

IPM also relies on the well learned features of each class for best performance in sample selection and thus must be performed at the conclusion of a task. This also makes IPM most suitable for the offline continual learning scenario.

Similar to herding, we again store $|\mathcal{M}|/s$ samples per class for a balanced and saturated buffer and preserve the most informative samples, as determined by IPM, by deleting the most recently added samples to each class when new data is encountered and must be stored to the fixed buffer.

\subsection{Dynamic Memory Buffers} \label{sec:dynamic_buffers}

Traditionally, when replay methods are used, it is common for the buffer size to be fixed \cite{ER, ER-ACE, DER}. This fixed buffer size does not take into account the underlying complexity of the data and more specifically, each class within the data. To account for such dataset specific complexities, we allow the buffer to be dynamic, where we add $K$ samples of class $c$ to the buffer using two algorithm agnostic \footnote{Algorithm agnostic refers to each population strategy as each strategy will choose different subsets of samples bounded by each studied dynamic buffer criterion.} methods we refer to as intracluster variance and Kaiser criterion described below. This idea is motivated by providing a guideline for determining the number of samples needed to represent the data manifold and the different classes within it as opposed to arbitrarily choosing a fixed buffer size.

\textbf{Intracluster Variance.} We assume that the number of images necessary for replay depends on the complexity of the underlying manifold of the data.  It is well known that relations between the data points on the high dimensional manifold are preserved when the data is embedded into a lower dimension via norm preserving transformations \cite{manifold_algorithms, testing_manifold}. For example, the dominant eigenvectors (or principal components) of the space in which the images lie is an example of one such transformation. In fact, the number of eigenvectors needed to represent the data is an indication of the complexity of the manifold, and therefore provides some indication of the number of images required for replay.  Another method for estimating the complexity of the manifold is by forming clusters in the data and by observing the change in the average intracluster variance for each cluster.

\begin{figure}[t]
  \centering
  \begin{subfigure}{0.49\linewidth}
    \includegraphics[width=1.0\linewidth]{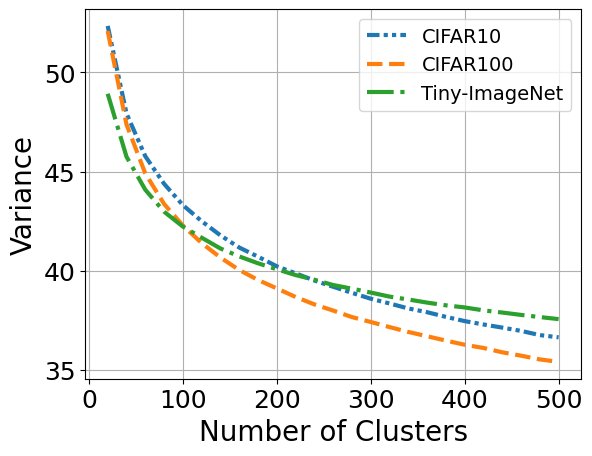}  
    \caption{}
    \label{fig:variance}
  \end{subfigure}
  \hfill
  \begin{subfigure}{0.49\linewidth}
    \includegraphics[width=1.0\linewidth]{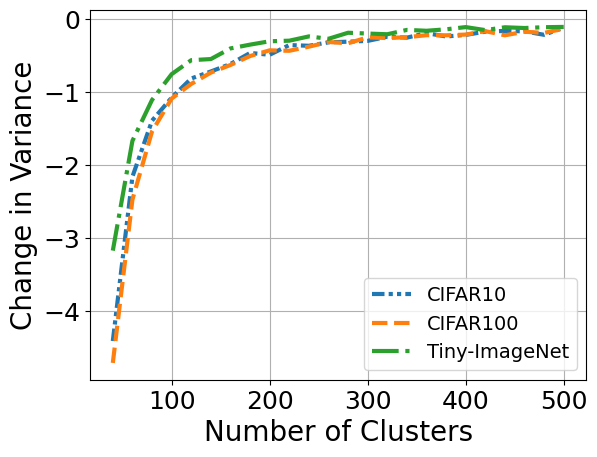}  
    \caption{}
    \label{fig:delta_variance}
  \end{subfigure}
  \caption{(a) Average intracluster variance for varying number of clusters. (b) Average change in intracluster variance for varying number of clusters. }
  \label{fig:Variance_Plots}
\end{figure}

Assume that the data is grouped into $K$ clusters, each with $N_j$ images. Specifically, let $x_{i,j}$ for $1 \leq i \leq N_j$, $1 \leq j \leq K$ represent the $i^{th}$ training image of the $j^{th}$ cluster. The variance of any given cluster is given by $\sigma_j^2 = \frac{1}{N_j} \sum_{i=1}^{N_j} (x_{i,j} - \mu_j)$ where $\mu_j = \frac{1}{N_j} \sum_{i=1}^{N_j} x_{i,j}$ is the mean of the respective cluster. The average variance of all clusters is then simply $\sigma_K^2 = \frac{1}{K} \sum_{j=1}^K \sigma_j^2$. The premise is that each cluster represents a local region on the manifold where the data is concentrated. As $K$ increases, each cluster becomes more and more compact and their variance $\sigma_j^2$ for $1 \leq j \leq K$ decreases. This causes the average variance $\sigma_K^2$ to also decrease as shown in Figure \ref{fig:variance} using the CIFAR10, CIFAR100 and Tiny-ImageNet datasets \cite{CIFAR, TIN} and the $k$-means clustering algorithm. In fact, the change in variance $\Delta_K = \sigma_{K+1}^2 - \sigma_K^2$ also tapers asymptotically as $K$ is allowed to increase. Figure \ref{fig:delta_variance} shows this behavior and the rate at which the average variance decreases.

We observe that increasing the number of clusters beyond the knee of the curve, 100 for both CIFAR10 and CIFAR100 and 160 for Tiny-ImageNet, provides diminishing gains in terms of decreasing the variance of each cluster. We assume that each cluster is as compact as possible, and the cluster mean is a good representation of the data that lies within the cluster. Therefore, we choose the number of clusters where the knee of the curve (found using the Kneedle algorithm~\cite{Kneedle}) in \ref{fig:variance} and \ref{fig:delta_variance} occur as an indication of the number of representative samples that are required to adequately represent the data manifold.

With this estimate of the number of samples needed for proper representation of the underlying data manifold, we populate the buffer with $K$ samples per class. This allows for the buffer to maintain its class balanced properly and avoid biased sampling from the buffer when training.

When using instracluster variance, one must have access to the entire dataset before training begins. This can be seen as unfair in the context of continual learning since in real world scenarios it would be impossible to view the entire dataset beforehand. We address this concern by using the Kaiser criterion below.

\textbf{Kaiser Criterion.} Instead of finding a global number of samples to be kept for each class as done in intracluster variance, we detail an additional method based on the number of most useful eigenvectors for each class.

Assume that $\mathbf{x}_{i,c}$ is a $d$ dimensional column vector that represents the $i^{th}$ training image of class $c$ for $1 \leq i \leq N_c$. For each class, we define the data matrix $\boldsymbol{A}_c = [\mathbf{x}_{1,c}, ..., \mathbf{x}_{1, N_c}]$, and compute the eigenvalues and eigenvectors of $\boldsymbol{A}_c\boldsymbol{A}_c^T$ denoted as $\lambda_{k,c}$ and $\boldsymbol\phi_{k,c}$, respectively, for $k=1,...,d$. Loosely speaking, the rank of this matrix is a proxy for the number of independent images required to represent the dataset and can thus be viewed as a reduced version of PCA. This in turn is related to the non-singular eigenvalues. Therefore, to determine the minimum number of images required to represent the data, we count the number of non-trivial eigenvalues (ignoring the ones that are zero or close to it). The Kaiser criterion \cite{kaiser_criterion} is a common method for selecting the useful eigenvectors which states that only those with eigenvalues greater than 1.0 should be retained for representing the data.

\begin{table}[t]
    \begin{center}
        \begin{tabular}{lccc}
        \toprule
        \textbf{Dataset} & \textbf{Mean} & \textbf{Min.} & \textbf{Max} \\
        \hline
        CIFAR10 & $248 \pm 27$ & 215 & 290 \\
        CIFAR100 & $50 \pm 8$ & 28 & 69 \\
        TinyImageNet & $59 \pm 9$ & 41 & 94 \\
        \bottomrule
        \end{tabular}
    \end{center}
\caption{Kaiser Criterion stats for number of samples required for each class.}
\label{Tab:kaiser_stats}
\end{table}

We note that this method of using raw images to form the data matrix is better suited than using network features, as that would always require the network to accurately classify the entire data, which manot not occur if a particular class is difficult for the network to classify.

We report the statistics of the Kaiser criterion for each dataset in Table \ref{Tab:kaiser_stats}. The benefits of using a dynamic buffer with the Kaiser criterion allow for classes with higher complexity to have more representation in the buffer, which ultimately leads to higher probability of sampling these more complex classes from the buffer and further mitigate forgetting.

Additionally, the Kaiser criterion can be used in an offline manner before the training of each task as the determination for the number of samples to be stored for each class depends only on a specific class's data matrix. Thus, when new task data is available, we can compute the Kaiser criterion for each individual class by partitioning the task data matrix into class data matrices.

\begin{table*}[htbp]
\small
\centering
\scriptsize
\begin{tabular}{@{}ccccccccc@{}c}
    \toprule
    & & & \multicolumn{2}{@{}c}{\textbf{Split-CIFAR10}} & \multicolumn{2}{@{}c}{\textbf{Split-CIFAR100}} & \multicolumn{2}{c}{\textbf{Split-TinyImageNet}} \\
    
    \textbf{\thead{Fixed\\ Buffer\\ Size}} & \textbf{Method} & \thead{\textbf{Population}\\ \textbf{Strategy}} & \multicolumn{2}{@{}c}{Class-IL} & \multicolumn{2}{@{}c}{Class-IL} & \multicolumn{2}{c}{Class-IL} \\
    
    & & & \emph{FAA} & \emph{FF} & \emph{FAA} & \emph{FF} & \emph{FAA} & \emph{FF} \\
    \midrule

    \multirow{16}{*}{200} & \multirow{4}{*}{ER} & Reservoir & $48.39 \pm 2.01$ & $60.39 \pm 2.26$ & $15.35 \pm 0.86$ & $81.20 \pm 0.62$ & $8.40 \pm 0.16$ & $76.88 \pm 0.18$ \\
    & & Herding & $\mathbf{52.32 \pm 0.77}$ & $\mathbf{55.17 \pm 0.77}$ & $\mathbf{15.94 \pm 0.46}$ & $\mathbf{79.60 \pm 0.11}$ & $\mathbf{8.81 \pm 0.05}$ & $\mathbf{76.61 \pm 0.62}$ \\
    & & GSS & $41.45 \pm 4.65$ & $68.92 \pm 5.57$ & $11.87 \pm 0.05$ & $84.34 \pm 0.41$ & - & - \\
    & & IPM & $48.68 \pm 1.11$ & $59.44 \pm 1.32$ & $15.0 \pm 0.26$ & $80.67 \pm 0.31$ & $8.55 \pm 0.09$ & $77.13 \pm 0.44$ \\
    \cline{2-9}
    
    & \multirow{4}{*}{DER} & Reservoir & $\mathbf{61.17 \pm 1.08}$ & $41.27 \pm 0.53$ & $24.38 \pm 1.46$ & $70.07 \pm 1.76$ & $11.15 \pm 0.46$ & $74.22 \pm 0.65$ \\
    & & Herding & $30.01 \pm 1.87$ & $83.12 \pm 2.01$ & $9.99 \pm 0.23$ & $87.73 \pm 0.50$ & $5.7 \pm 1.06$ & $71.92 \pm 1.87$ \\
    & & GSS & $38.04 \pm 6.09$ & $71.95 \pm 7.60$ & $12.29 \pm 1.70$ & $78.05 \pm 0.53$ & - & - \\
    & & IPM & $60.48 \pm 0.34$ & $\mathbf{30.12 \pm 1.03}$ & $\mathbf{33.47 \pm 1.86}$ & $\mathbf{42.32 \pm 1.71}$ & $\mathbf{19.36 \pm 1.06}$ & $\mathbf{45.46 \pm 0.63}$ \\
    \cline{2-9}
    
    & \multirow{4}{*}{GDumb} & Reservoir & $29.26 \pm 1.18$ & N/A & $4.63 \pm 0.49$ & N/A & $2.13 \pm 0.28$ & N/A\\
    & & Herding & $\mathbf{32.16 \pm 1.51}$ & N/A & $\mathbf{7.02 \pm 0.51}$ & N/A & $\mathbf{3.45 \pm 0.30}$ & N/A \\
    & & GSS & $28.35 \pm 1.19$ & N/A & $4.81 \pm 0.29$ & N/A & - & N/A \\
    & & IPM & $31.60 \pm 1.83$ & N/A & $6.26 \pm 0.14$ & N/A & $2.67 \pm 0.36$ & N/A \\
    \cline{2-9}
    
    & \multirow{4}{*}{ER-ACE} & Reservoir & $\mathbf{63.32 \pm 2.40}$ & $18.62 \pm 2.24$ & $28.78 \pm 0.66$ & $44.40 \pm 1.13$ & $12.82 \pm 0.11$ & $48.91 \pm 1.65$ \\
    & & Herding & $61.66 \pm 0.35$ & $33.04 \pm 3.37$ & $\mathbf{29.64 \pm 0.29}$ & $62.08 \pm 0.27$ & $\mathbf{15.75 \pm 0.09}$ & $64.86 \pm 1.45$ \\
    & & GSS & $26.25 \pm 9.53$ & $\mathbf{1.02 \pm 1.01}$ & $7.525 \pm 0.13$ & $\mathbf{8.93 \pm 0.75}$ & - & - \\
    & & IPM & $49.41 \pm 0.56$ & $16.56 \pm 0.09$ & $28.36 \pm 0.28$ & $27.17 \pm 0.23$ & $15.02 \pm 0.53$ & $\mathbf{29.16 \pm 0.35}$ \\

    \hline
    \multirow{16}{*}{500} & \multirow{4}{*}{ER} & Reservoir & $61.07 \pm 0.64$ & $44.27 \pm 1.10$ & $21.37 \pm 1.27$ & $73.55 \pm 0.77$ & $10.19 \pm 0.20$ & $75.34 \pm 0.16$ \\
    & & Herding & $\mathbf{64.09 \pm 0.73}$ & $\mathbf{39.26 \pm 1.39}$ & $\mathbf{24.25 \pm 0.70}$ & $\mathbf{70.46 \pm 0.57}$ & $\mathbf{10.38 \pm 0.16}$ & $75.50 \pm 0.35$ \\
    & & GSS & $61.07 \pm 0.64$ & $44.27 \pm 1.10$ & $21.37 \pm 1.27$ & $73.55 \pm 0.77$ & - & - \\
    & & IPM & $61.08 \pm 0.46$ & $44.33 \pm 0.68$ & $22.06 \pm 0.48$ & $72.80 \pm 0.33$ & $10.20 \pm 0.17$ & $\mathbf{74.98 \pm 0.18}$ \\
    \cline{2-9}
    
    & \multirow{4}{*}{DER} & Reservoir & $\mathbf{70.07 \pm 0.95}$ & $29.5 \pm 1.80$ & $34.53 \pm 1.68$ & $56.30 \pm 1.52$ & $17.15 \pm 1.40$ & $66.91 \pm 1.81$ \\
    & & Herding & $48.20 \pm 2.94$ & $54.64 \pm 7.93$ & $13.11 \pm 0.51$ & $84.04 \pm 0.69$ & $5.32 \pm 0.78$ & $67.13 \pm 2.02$ \\
    & & GSS & $45.94 \pm 6.22$ & $59.09 \pm 9.16$ & $16.64 \pm 1.69$ & $72.323 \pm 5.59$ & - & - \\
    & & IPM & $65.5 \pm 2.68$ & $\mathbf{23.55 \pm 5.19}$ & $\mathbf{40.51 \pm 0.43}$ & $\mathbf{28.97 \pm 1.06}$ & $\mathbf{20.49 \pm 0.86}$ & $\mathbf{31.54 \pm 2.61}$ \\
    \cline{2-9}

    & \multirow{4}{*}{GDumb} & Reservoir & $\mathbf{43.35 \pm 0.55}$ & N/A & $9.85 \pm 0.45$ & N/A & $3.6 \pm 0.004$ & N/A \\
    & & Herding & $42.85 \pm 0.83$ & N/A & $\mathbf{11.45 \pm 0.42}$ & N/A & $\mathbf{4.83 \pm 0.19}$ & N/A \\
    & & GSS & $37.39 \pm 1.21$ & N/A & $6.2 \pm 0.29$ & N/A & - & N/A \\
    & & IPM & $42.03 \pm 3.05$ & N/A & $9.02 \pm 0.67$ & N/A & $3.27 \pm 0.40$ & N/A \\
    \cline{2-9}

    & \multirow{4}{*}{ER-ACE} & Reservoir & $\mathbf{72.15 \pm 0.38}$ & $13.18 \pm 1.26$ & $\mathbf{37.60 \pm 0.15}$ & $38.17 \pm 1.16$ & $\mathbf{20.99 \pm 0.52}$ & $46.60 \pm 0.55$ \\
    & & Herding & $69.75 \pm 1.90$ & $23.00 \pm 4.32$ & $37.54 \pm 0.44$ & $53.08 \pm 0.47$ & $20.3 \pm 0.36$ & $60.48 \pm 0.06$ \\
    & & GSS & $19.54 \pm 0.09$ & $\mathbf{0.22 \pm 0.11}$ & $8.34 \pm 0.09$ & $\mathbf{7.73 \pm 0.62}$ & - & - \\
    & & IPM & $54.43 \pm 0.99$ & $14.57 \pm 1.17$ & $35.01 \pm 0.14$ & $25.50 \pm 0.91$ & $18.5 \pm 0.42$ & $\mathbf{34.41 \pm 8.08}$ \\

    \hline
    \multirow{16}{*}{5120} & \multirow{4}{*}{ER} & Reservoir & $83.18 \pm 1.82$ & $14.33 \pm 1.65$ & $50.71 \pm 0.27$ & $38.92 \pm 0.44$ & $27.36 \pm 0.03$ & $54.46 \pm 0.69$ \\
    & & Herding & $\mathbf{85.56 \pm 0.34}$ & $12.22 \pm 0.29$ & $\mathbf{52.93 \pm 1.55}$ & $\mathbf{35.21 \pm 0.93}$ & $\mathbf{28.43 \pm 0.51}$ & $\mathbf{52.12 \pm 0.51}$ \\
    & & GSS & $60.35 \pm 7.06$ & $43.74 \pm 8.69$ & $17.52 \pm 0.22$ & $78.74 \pm 2.28$ & - & - \\
    & & IPM & $85.19 \pm 0.63$ & $\mathbf{10.97 \pm 0.91}$ & $50.75 \pm 1.16$ & $36.96 \pm 0.60$ & $27.39 \pm 0.23$ & $53.43 \pm 0.12$ \\
    \cline{2-9}
    
    & \multirow{4}{*}{DER} & Reservoir & $\mathbf{83.35 \pm 0.72}$ & $11.27 \pm 0.96$ & $57.22 \pm 0.24$ & $22.86 \pm 1.51$ & $\mathbf{37.09 \pm 0.50}$ & $31.91 \pm 1.05$ \\
    & & Herding & $76.21 \pm 1.08$ & $25.26 \pm 1.63$ & $52.53 \pm 0.24$ & $38.70 \pm 0.38$ & $5.09 \pm 2.59$ & $38.44 \pm 5.37$ \\
    & & GSS & $45.86 \pm 16.34$ & $54.54 \pm 21.02$ & $56.32 \pm 1.10$ & $45.05 \pm 7.04$ & - & - \\
    & & IPM & $67.75 \pm 0.52$ & $\mathbf{6.25 \pm 1.21}$ & $\mathbf{57.30 \pm 0.51}$ & $\mathbf{8.85 \pm 0.52}$ & $34.33 \pm 0.93$ & $\mathbf{7.27 \pm 0.98}$ \\
    \cline{2-9}
    
    & \multirow{4}{*}{GDumb} & Reservoir & $\mathbf{79.89 \pm 1.29}$ & N/A & $\mathbf{42.52 \pm 0.22}$ & N/A & $\mathbf{21.18 \pm 0.06}$ & N/A \\
    & & Herding & $77.16 \pm 0.74$ & N/A & $36.80 \pm 0.57$ & N/A & $17.38 \pm 0.40$ & N/A \\
    & & GSS & $70.27 \pm 2.29$ & N/A & $19.64 \pm 1.40$ & N/A & - & N/A \\
    & & IPM & $79.58 \pm 0.93$ & N/A & $42.31 \pm 0.18$ & N/A & $20.99 \pm 0.72$ & N/A \\
    \cline{2-9}
    
    & \multirow{4}{*}{ER-ACE} & Reservoir & $83.67 \pm 0.26$ & $4.93 \pm 0.30$ & $57.01 \pm 0.27$ & $21.02 \pm 0.22$ & $\mathbf{38.68 \pm 0.37}$ & $29.41 \pm 0.74$ \\
    & & Herding & $\mathbf{85.02 \pm 0.86}$ & $11.83 \pm 2.08$ & $\mathbf{58.52 \pm 0.23}$ & $26.77 \pm 0.19$ & $34.66 \pm 0.99$ & $41.91 \pm 1.08$ \\
    & & GSS & $19.73 \pm 0.01$ & $\mathbf{0.05 \pm 0.02}$ & $9.20 \pm 0.04$ & $\mathbf{4.31 \pm 0.71}$ & - & - \\
    & & IPM & $66.69 \pm 0.46$ & $6.29 \pm 0.38$ & $53.91 \pm 0.43$ & $14.69 \pm 0.46$ & $36.28 \pm 0.19$ & $\mathbf{18.99 \pm 0.27}$ \\
     
    \bottomrule
  \end{tabular}
  \caption{Population strategy results tested with various replay based methods with traditionally used fixed size buffer, averaged across three runs. We do not report forgetting in GDumb experiments due to the nature of GDumb only training on the fully populated, balanced buffer. Results for TinyImageNet are not reported for GSS due to intractable train times.}
  \label{Tab:fixed_buffer_results}
\end{table*}

\section{Experiments}

We investigate the performance of each population strategy described above by comparing each scheme under a commonly used suite of replay-based training methods. We test using both a fixed memory buffer with commonly used buffer sizes and the newly proposed dynamic buffer scheme described in Section \ref{sec:dynamic_buffers}. Our primary focus is on the offline class-IL setting. We report offline task-IL results in the supplementary material.

\textbf{Datasets.} We benchmark each population strategy on three commonly used continual learning datasets: split-CIFAR10, split-CIFAR100 and split-Tiny-ImageNet~\cite{CIFAR, splitCIFAR, TIN}. In split-CIFAR10, the CIFAR10 dataset is split into 5 disjoint tasks where each task contains 2 classes. Split-CIFAR100 splits CIFAR100 into 10 disjoint tasks of 10 classes each. Split-Tiny-ImageNet splits Tiny-ImageNet into 10 disjoint tasks of 20 classes each. We maintain the same order in which classes are split across all tested population strategies and methods.

\textbf{Compared Methods.} Each population strategy studied herein is tested against four commonly compared methods in continual learning literature, namely ER \cite{ER}, DER \cite{DER}, GDumb \cite{GDumb}, and ER-ACE \cite{ER-ACE}. Each of the aforementioned methods was state of the art for its time with ER-ACE being the most recently proposed state of the art method for bench-marking of current replay-based continual learning research.

All compared methods use some form of experience replay. For each $t>1$, all methods train on the union of task data and data stored in memory as $\mathcal{D}_t \cup \mathcal{M}$, except in the case of GDumb where we only populate $\mathcal{M}$ in a balanced manner by performing one iteration through each $\mathcal{D}_t$ and proceed to train solely on $\mathcal{M}$ at the conclusion of the final task.

\textbf{Configuration and Hyperparameters.} We test each of the buffer population strategies studied herein with the above described methods using the open-source codebase, \emph{Mammoth}, first introduced in \cite{DER}. We use the best configuration for each compared training method when testing performance of each population strategy with a ResNet18 \cite{ResNet} backbone for fair comparisons. Exact hyperparmeter configuration can be found in the supplementary material.

\begin{figure*}
  \centering
  
  \begin{subfigure}{0.33\linewidth}
    \includegraphics[width=1.0\linewidth]{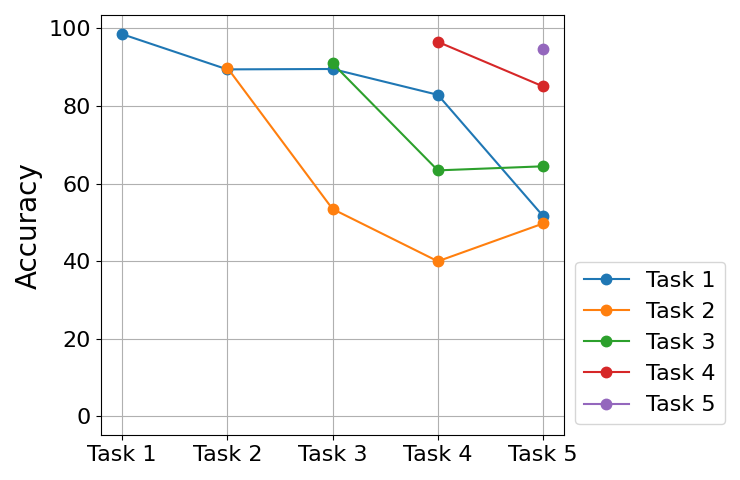}  
    \caption{}
    \label{fig:cifar10_original_500}
  \end{subfigure}
  \begin{subfigure}{0.33\linewidth}
    \includegraphics[width=1.0\linewidth]{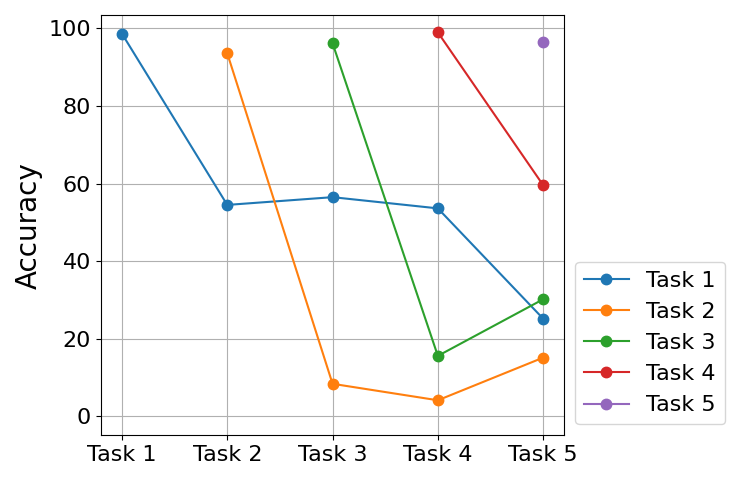}  
    \caption{}
    \label{fig:cifar10_herding_500}
  \end{subfigure}
  \begin{subfigure}{0.33\linewidth}
    \includegraphics[width=1.0\linewidth]{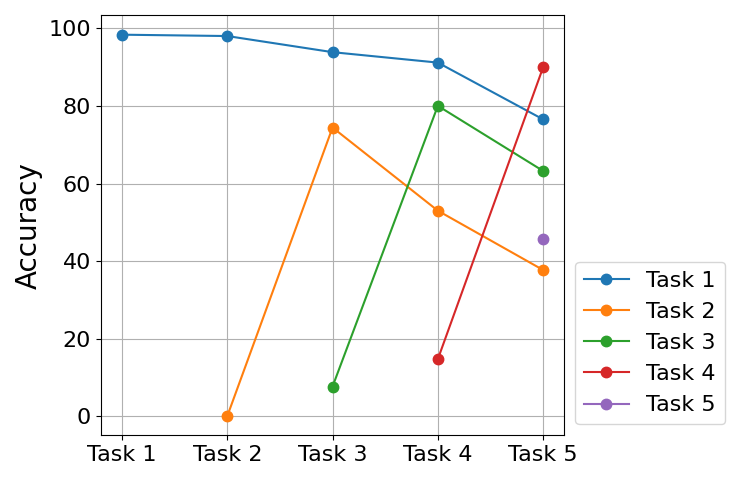}   
    \caption{}
    \label{fig:cifar10_ipm_500}
  \end{subfigure}
  \begin{subfigure}{0.33\linewidth}
    \includegraphics[width=1.0\linewidth]{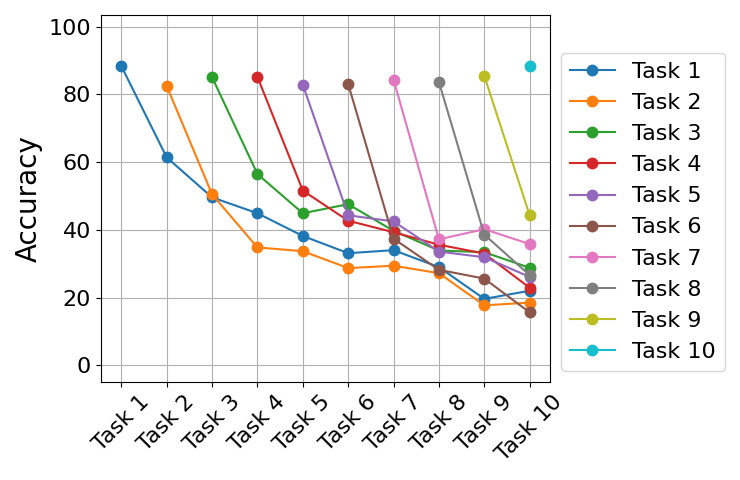}  
    \caption{}
    \label{fig:cifar100_original_500}
  \end{subfigure}
  \begin{subfigure}{0.33\linewidth}
    \includegraphics[width=1.0\linewidth]{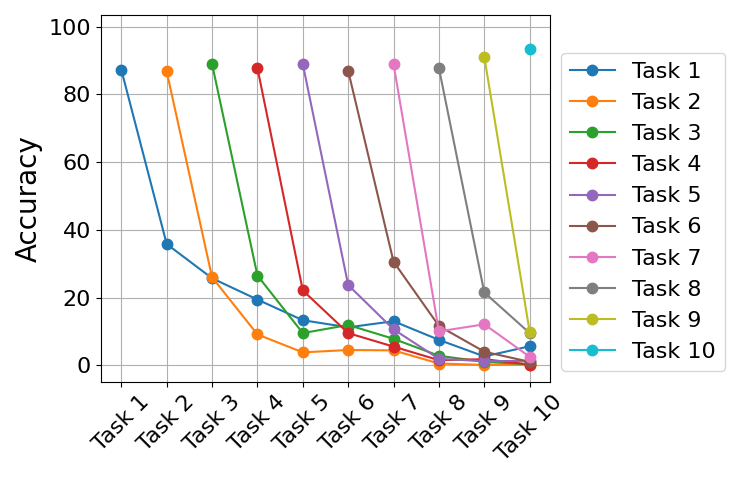}  
    \caption{}
    \label{fig:cifar100_herding_500}
  \end{subfigure}
  \begin{subfigure}{0.33\linewidth}
    \includegraphics[width=1.0\linewidth]{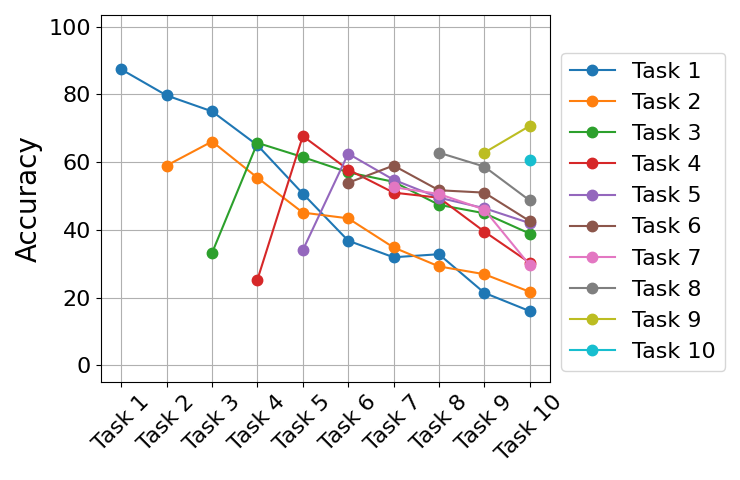}   
    \caption{}
    \label{fig:cifar100_ipm_500}
  \end{subfigure}
  \caption{A comparison of the reservoir, herding, and IPM population strategies paired with DER with a fixed buffer size of 500. Top row corresponds to Split-CIFAR10 performance and bottom row is Split-CIFAR100. The columns correspond as follows: left uses reservoir sampling, center uses herding, and right uses IPM. We do not report GSS results due to all around inferior performance.}
  \label{fig:population_comparissons}
\end{figure*}

\textbf{Metrics.} We judge performance with the commonly used metrics of final average accuracy (\emph{FAA}) and final forgetting (\emph{FF}) given by
\begin{equation}
    \label{FAA}
        FAA = \frac{1}{\mathcal{T}} \sum_{j=1}^{\mathcal{T}} a_j^{\mathcal{T}} \\
\end{equation}
\begin{equation}
    \label{FF}
        FF = \frac{1}{\mathcal{T}-1} \sum_{j=1}^{\mathcal{T}-1} f_j^{\mathcal{T}} \text{ s.t. } f_j^{\mathcal{T}} = \underset{l \in \{1,...,\mathcal{T}-1\}}{\operatorname{max}} a_l^l - a_j^{\mathcal{T}}\\
\end{equation}
where $a_j^{\mathcal{T}}$ and $f_j^{\mathcal{T}}$ are interpreted as the accuracy and forgetting of task $j$ and the end of training on $\mathcal{T}$ tasks respectively \cite{riemann_walk}. When judging performance, we seek maximal \emph{FAA} and minimal \emph{FF}.

\begin{table}[h]
    \begin{center}
        \begin{tabular}{lc}
        \toprule
        \textbf{\thead{Fixed\\ Buffer\\ Size}} & \textbf{\thead{Percentage of Samples\\ Belonging to Each Task\\ ($t_1/t_2/t_3/t_4/t_5$)}} \\
        \hline
        200 & 23.5\% / 19.0\% / 16.0\% / 22.0\% / 19.5\% \\
        500 & 23.5\% / 15.0\% / 20.5\% / 22.5\% / 18.5\% \\
        5120 & 20.0\% / 18.5\% / 19.53\% / 20.43\% / 21.54\% \\
        \bottomrule
        \end{tabular}
    \end{center}
    \caption{Percentage of samples in buffer belonging to each task at the end of training populated via reservoir sampling.}
    \label{Tab:reservoir_buffer_imbalance}
\end{table}

\subsection{Results}

\textbf{Fixed Buffer.} Results for commonly tested fixed buffer sizes are in Table \ref{Tab:fixed_buffer_results}. We first observe that in nearly all cases, reservoir sampling leads to greater forgetting when compared to the other population strategies, specifically compared to herding and IPM.  We also observe in several trials, that reservoir sampling also underperforms in \emph{FAA} compared to herding and IPM. This behavior can be attributed to both herding and IPM selecting the best samples from the learned feature space at the conclusion of each task whereas reservoir sampling simply selects and replaces samples  at random. 

Another reason for greater forgetting in reservoir sampling is the unbalanced fixed buffers incurred by the random sampling and replacement of data in the buffer. We show the percentage of task specific data of the final fixed buffer when populated via reservoir sampling in Table \ref{Tab:reservoir_buffer_imbalance}. We can clearly see that, for smaller buffer sizes, there is a bias to the storage of earlier task data compared to more recently encountered tasks. In comparison, both herding and IPM maintain a balanced fixed buffer at all times leading to equal probability of sampling any task data for batch training. Naturally, as the buffer size increases, the unbalanced nature of the buffer populated with reservoir sampling becomes less severe and we see the reservoir population strategy become more competitive with herding and IPM in both \emph{FAA} and \emph{FF}.

\begin{figure*}[th]
  \centering
  \begin{subfigure}{0.33\linewidth}
    \includegraphics[width=1.0\linewidth]{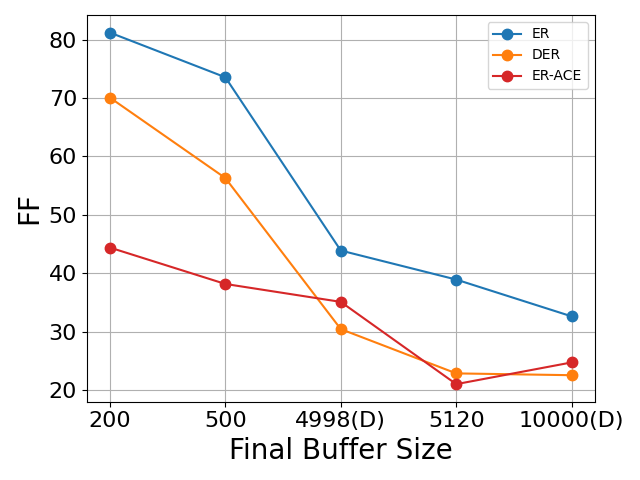}  
    \caption{}
    \label{fig:cifar10_dynamic}
  \end{subfigure}
  \begin{subfigure}{0.33\linewidth}
    \includegraphics[width=1.0\linewidth]{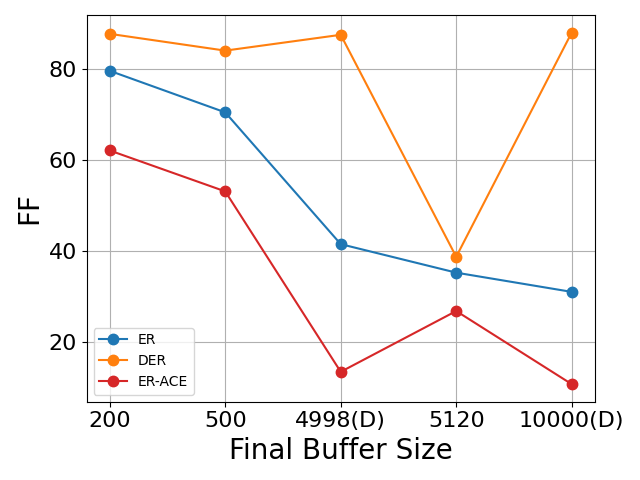}  
    \caption{}
    \label{fig:cifar10_dynamic}
  \end{subfigure}
  \begin{subfigure}{0.33\linewidth}
    \includegraphics[width=1.0\linewidth]{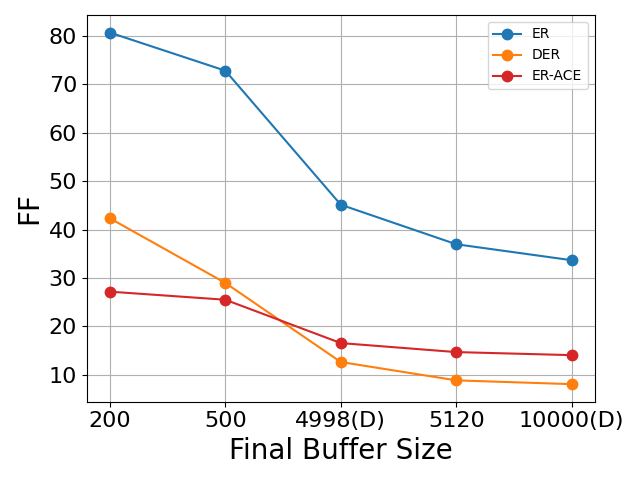}   
    \caption{}
    \label{fig:cifar10_dynamic}
  \end{subfigure}
  \caption{Final forgetting performance with various final buffer sizes tested with Split-CIFAR100. Final buffer sizes with a (D) indicate dynamic final size. In order from left to right are results from reservoir, herding, and IPM respectively.}
  \label{fig:dyanmic_vs_fixed_FF}
\end{figure*}

We pay particular interest to the scenarios where IPM yields superior \emph{FF} yet inferior \emph{FAA}. To analyze why this happens in certain cases, we plot each population strategy used in conjunction with DER for both split-CIFAR10 and split-CIFAR100 in Figure \ref{fig:population_comparissons} (we omit GSS figures due to all around inferior performance). We observe the interesting behavior where for each $t>1$, IPM initially has poor current task performance but then proceeds to make astonishing recoveries for each subsequent task. This indicates that IPM tends to prioritize performance on the buffer instead of the current task at hand which in turn leads to lesser forgetting. An interesting observation to make is that this behavior holds only for DER and ER-ACE (see supplementary material for additional figures). Both DER and ER-ACE are training schemes that directly optimize on logits indicating that IPM is best suited for these types of training schemes.

\begin{figure}[t]
  \centering
  \begin{subfigure}{0.49\linewidth}
    \includegraphics[width=1.0\linewidth]{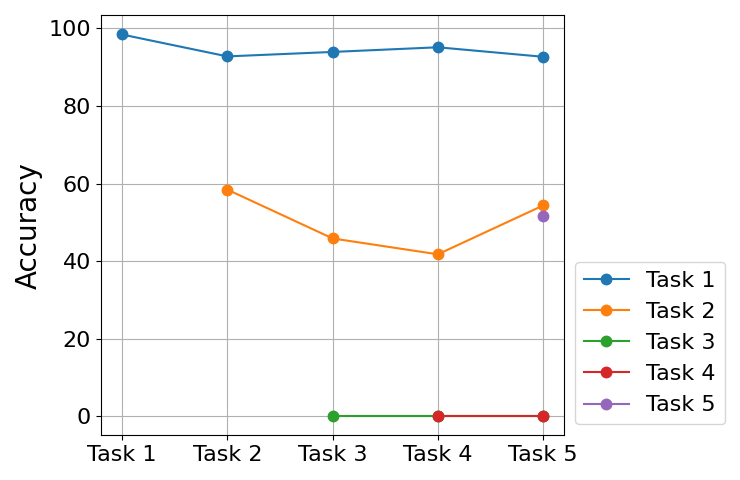}  
    \caption{}
    \label{fig:gss_cifar10}
  \end{subfigure}
  \hfill
  \begin{subfigure}{0.49\linewidth}
    \includegraphics[width=1.0\linewidth]{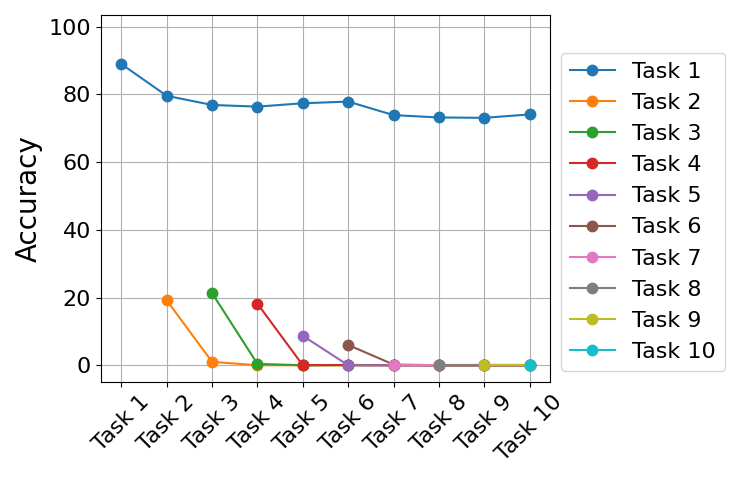}  
    \caption{}
    \label{fig:gss_cifar100}
  \end{subfigure}
  \caption{ER-ACE using the GSS population strategy for a fixed buffer with buffer size 200. (a) Split-CIFAR10 results. (b) Split-CIFAR100 results.}
  \label{fig:gss_er-ace}
\end{figure}

We take note of the relatively low forgetting of the GSS population strategy when tested with ER-ACE. To investigate why GSS achieves such low forgetting, we plot the accuracy of each previously learned task as new tasks are learned in Figure \ref{fig:gss_er-ace}. From this, we can infer that GSS's low forgetting capabilities when coupled with ER-ACE is caused by the inferior performance on subsequent tasks after $t=1$ and thus, has nearly nothing to forget. We note the strong performance of $t=1$ throughout the model's life, however. This is due to GSS first populating the buffer with $t=1$ data and thereafter, hardly ever finding a sample with an appropriate score to replace other samples in the buffer as described in Section \ref{sec: GSS_sec}. Because of this nature, we should not accept that GSS is the best forgetting performer with coupled with ER-ACE. We still observe that reservoir sampling mostly does not compare in forgetting to other strategies when ignoring GSS results.

We observe the all around competitiveness of each population strategy when tested with GDumb (note, we do not report \emph{FF} for GDumb as there is no forgetting to take place since GDumb trains solely on the buffer). Because GDumb only uses each observed task to populate the memory buffer, it makes sense that herding and IPM perform roughly the same with reservoir sampling since herding and IPM depend on the well learned features for population. Similarly, GSS depends on the gradients of each sample, but because GDumb takes no gradient steps until the conclusion of the final observed task, GSS has no proper way to score samples for replacement or not.

Lastly, we make the note that while no single method consistently outperforms any other, we demonstrate that in many situations, reservoir sampling yields inferior final forgetting performance. This suggests that when using replay-based methods in continual learning solutions, reservoir sampling should not be blindly used as the buffer population strategy of choice, and one should instead pay careful attention to selection of buffer population algorithm for the best performance.

\textbf{Dynamic Buffer.} We next perform experiments using dynamic buffers using the two criteria as described in Section \ref{sec:dynamic_buffers} and provide tabulated results for class-IL and task-IL scenarios in the supplementary material. We omit GSS results using a dynamic buffer due to poor performance with paired with any of the fixed buffer schemes.

Overall, we observe much of the same trends as seen with fixed buffers. However, we notice that dynamic buffers seem to benefit most when paired with reservoir sampling and IPM in Figure \ref{fig:dyanmic_vs_fixed_FF} (\emph{FAA} and \emph{FF} for all datasets are reported in supplementary material). Because dynamic buffers find the optimal number of samples for storage, we expect the change in \emph{FF} to be lower when we approach that number, which we observe in Figure \ref{fig:dyanmic_vs_fixed_FF}. In general we observe the Kaiser criterion performing better than intracluster variance. This can be attributed to the ability to adapt to classs complexity for the Kaiser criterion, particularly as the number of classes increases in a dataset.

We note the curious performance of DER when coupled with dynamic replay and give a brief conjecture for why this may be in the supplemental material.

\section{Conclusions}
In this work, we compare the commonly used approach of reservoir sampling for memory buffer population in replay methods to other greedy sampling methods to answer the question of \emph{which samples should be stored} in memory. We show that reservoir sampling tends to lead to higher forgetting when compared to methods that use strategic population strategies to select the best data points for storage. We then address the question of \emph{how many samples should be stored} by the formulation of a dynamic buffer populated according to two criteria based on a dataset's eigenvectors and eigenvalues. We show that dynamic buffers lead to more competitive performance for all population strategies when compared to the arbitrary fixed buffer sizes commonly used in replay methods.

{\small
\bibliographystyle{ieee_fullname}
\bibliography{main_references}
}

\end{document}